%% file: iclr2020_conference.tex
\documentclass{article} % For LaTeX2e
\usepackage{iclr2020_conference,times}

\input{math_commands.tex}

\usepackage{url}
\usepackage{color}
\usepackage{xspace}
\usepackage{caption}
\usepackage{subcaption}
\usepackage{graphicx}
\usepackage{tabularx}
\usepackage{booktabs}
\usepackage{algorithm2e}

\definecolor{darkblue}{rgb}{0.0, 0.0, 0.5}
\usepackage[pagebackref=false,colorlinks=true,linkcolor=darkblue,citecolor=darkblue,bookmarks=false]{hyperref}

\DeclareCaptionLabelFormat{andtable}{#1~#2  \&  \tablename~\thetable}
\newcommand{\flops}{FLOPs\xspace}

\title{AtomNAS: Fine-Grained End-to-End Neural Architecture Search}

\author{Jieru Mei$^{1 {\ast}}$, Yingwei Li$^{1 {\ast}}$, Xiaochen Lian$^2$, Xiaojie Jin$^2$, Linjie Yang$^2$, \\\textbf{Alan Yuille$^1$} \textbf{\& Jianchao Yang$^2$} \\
$^1$Johns Hopkins University\\
$^2$ByteDance AI Lab\\
\texttt{\tiny meijieru@gmail.com, yingwei.li@jhu.edu, \{xiaochen.lian, jinxiaojie, linjie.yang\}@bytedance.com, } \\
\texttt{\tiny alan.l.yuille@gmail.com, yangjianchao@bytedance.com}
}

\iclrfinalcopy % Uncomment for camera-ready version, but NOT for submission.
\begin{document}

\maketitle

\begin{abstract}

Search space design is very critical to neural architecture search (NAS) algorithms. We propose a fine-grained search space comprised of atomic blocks, a minimal search unit that is much smaller than the ones used in recent NAS algorithms.
% \prev{This search space facilitates direct selection of channel numbers and kernel sizes in convolutions. In addition, we propose a resource-aware architecture search algorithm which dynamically selects atomic blocks during training. The algorithm is further accelerated by a dynamic network shrinkage technique. Instead of a search-and-retrain two-stage paradigm, our method simultaneously searches and trains the target architecture in an end-to-end manner.} 
This search space allows a mix of operations by composing different types of atomic blocks, while the search space in previous methods only allows homogeneous operations. Based on this search space, we propose a resource-aware architecture search framework which automatically assigns the computational resources (e.g., output channel numbers) for each operation by jointly considering the performance and the computational cost. In addition, to accelerate the search process, we propose a dynamic network shrinkage technique which prunes the atomic blocks with negligible influence on outputs on the fly. 
Instead of a search-and-retrain two-stage paradigm, our method simultaneously searches and trains the target architecture. 
Our method achieves state-of-the-art performance under several \flops configurations on ImageNet with a small searching cost.
We open our entire codebase at: \url{https://github.com/meijieru/AtomNAS}.

%We achieve $79.0\%$ top-1 accuracy under 600M \flops constraint on ImageNet, which is a new state-of-the-art under the typical mobile setting.
\end{abstract}

\let\thefootnote\relax\footnote{$^\ast$ This work was done during the internship program at Bytedance.}

\section{Introduction}

Human-designed neural networks are already surpassed by machine-designed ones. Neural Architecture Search (NAS) has become the mainstream approach to discover efficient and powerful network structures (\cite{zoph2017nasnet, hieu2018enas, tan2019mnasnet, hanxiao2019darts}). Although the tedious searching process is conducted by machines, humans still involve extensively in the design of the NAS algorithms. Designing of search spaces is critical for NAS algorithms and different choices have been explored. \cite{han2019proxyless} and~\cite{wu2019fbnet} utilize supernets with multiple choices in each layer to accommodate a sampled network on the GPU. \cite{chen2019pdarts} progressively grow the depth of the supernet and remove unnecessary blocks during the search. \cite{mingxing2019efficient} propose to search the scaling factor of image resolution, channel multiplier and layer numbers in scenarios with different computation budgets. \cite{stamoulis2019single_path} propose to use different kernel sizes in each layer of the supernet and reuse the weights of larger kernels for small kernels. ~\cite{howard2019mobilenetv3,tan2019mixnet} adopts Inverted Residuals with Linear Bottlenecks (MobileNetV2 block) \citep{sandler2018mobilenetv2}, a building block with light-weighted depth-wise convolutions for highly efficient networks in mobile scenarios. 
%Search space design is a very crucial part in automatic neural architecture search (NAS). Though having the potential to obtain the optimal architecture, large search space, e.g., \cite{zoph2017nasnet}, usually comes with a big computational burden, making it almost impossible to be fully explored by a search algorithm. 

However, the proposed search spaces generally have only a small set of choices for each block. DARTS and related methods~\citep{hanxiao2019darts,chen2019pdarts, liang2019dartsplus} use around 10 different operations between two network nodes. \cite{howard2019mobilenetv3,han2019proxyless,wu2019fbnet,stamoulis2019single_path} search the expansion ratios in the MobileNetV2 block but still limit them to a few discrete values. We argue that search space of finer granularity is critical to find optimal neural architectures. Specifically, the searched building block in a supernet should be as small as possible to generate the most diversified model structures.
%Among these methods, quite a few adopt the strategy that restricts the search within a set of effective basic blocks \citep{tan2019mnasnet, howard2019mobilenetv3, tan2019mixnet}. For example, \cite{tan2019mixnet} uses Inverted Residual with Linear Bottleneck plus Squeeze-and-Excite blocks as its search units and achieves $78.9\%$ top-1 accuracy on ImageNet \citep{deng2009imagenet} under the typical mobile setting (\flops$<$600M). As a compromise for efficiency, block-wise search paradigm often operates at a coarse level, e.g., the types and the configurations of the operations are either determined by heuristic rules (!!!cite!!!) or selected from a limited set of choices (!!!cite!!!). 

We revisit the architectures of state-of-the-art networks (\cite{howard2019mobilenetv3,tan2019mixnet,he2016resnet}) and discover a commonly used building structure: convolution - channel-wise operation - convolution. We reinterpret this building structure as an ensemble of computationally independent blocks, which we call \textit{atomic blocks}. 
%\prev{This new formulation enables a much larger and more fine-grained search space.}
As the minimum search unit, the atomic block constitutes a much larger and more fine-grained search space, within which we are able to search for mixed operations (e.g., convolutions with different kernel sizes and their channel numbers).
%\prev{Starting from a supernet which is built upon atomic blocks, the search for exact channel numbers and various operations can be achieved by selecting a subset of the atomic blocks.} 
%\new{By composing different kinds of atomic blocks, the channel number of each candidate operations varies.}
%\lxc{A sentence describing the supernet: within each building block, use mix operations and initial channel numbers. Starting from this supernet, ...}
%A close look at the architectures of state-of-the-art networks (\cite{howard2019mobilenetv3,tan2019mixnet,he2016resnet}) reveals a commonly used basic structure: two convolutions joined by a depth-wise operation. We provide a new perspective to interpret such structure as an ensemble of computationally independent blocks, which we call \textit{atomic blocks}. This new formulation enables a much larger and more fine-grained search space; starting from a supernet which is built upon atomic blocks, the search for channel numbers and mix operations are achieved by determining whether to keep each atomic block. Many network architectures are instances within this new search space.

For the efficient exploration of the new search space, we propose a NAS framework named AtomNAS which applies network pruning techniques to architecture search.
%Specifically, we start from a supernet whose building blocks use different kinds of atomic blocks with initial channel numbers, where an importance factor is introduced to each atomic block.} \prev{we propose a NAS algorithm named AtomNAS to conduct architecture search and network training simultaneously. Specifically, an importance factor is introduced to each atomic block.} A penalty term proportional to the computation cost of the atomic block is enforced on the network. \new{Our method conducts architecture search and network training simultaneously by jointly learning the importance factors along with the weights of the network.} \prev{By jointly learning the importance factors along with the weights of the network, AtomNAS}, \new{It} selects the atomic blocks which contribute to the model capacity with relatively small computation cost.
Specifically, we start from an initial large supernet and rewrite every convolution - channel-wise operation - convolution structure of it in the form the weighted sum of atomic blocks; the weights reflect the contribution of the atomic blocks to the network capacity and are called \textit{importance factors}. For each atomic block, a penalty term in proportion to its FLOPs is enforced on its importance factor; effectively, the penalty makes AtomNAS favor atomic blocks with less FLOPs. By minimizing the combination of the original network loss and the total penalty on the weights, AtomNAS is able to learn both the parameters of the network and the weights of the atomic blocks. At the end of the learning, atomic blocks with very small weights (e.g., $<0.001$) are removed from the network and we obtain the final network which has fewer FLOPs. Since the pruned atomic blocks have little contribution to the network output due to their negligible weights, the final network does not need to be retrained or finetuned.

Training on the large supernet is computationally demanding. We observe that for many pruned atomic blocks, their weights diminish at the early stage of learning and never ``revive" throughout the rest of learning. We propose a dynamic network shrinkage technique which removes those atomic blocks on the fly and greatly reduces the run time of AtomNAS.

In our experiment, our method achieves 75.9\% top-1 accuracy on ImageNet dataset around 360M \flops, which is 0.9\% higher than state-of-the-art model \citep{stamoulis2019single_path}. By further incorporating additional modules, our method achieves 77.6\% top-1 accuracy. It outperforms MixNet by 0.6\% using 363M \flops, which is a new state-of-the-art under the mobile scenario.

In summary, the major contributions of our work are:
\begin{enumerate}
    \item We design a fine-grained search space which includes the exact number of channels and mixed operations (e.g., combination of different convolution kernels).
    \item We propose an NAS framework, AtomNAS. Within the framework, an efficient end-to-end NAS algorithm is proposed which can simultaneously search the network architecture and train the final model. No finetuning is needed after the algorithm finishes.
    % \prev{\item We propose an efficient end-to-end NAS algorithm named AtomNAS which can simultaneously search the network architecture and train the final model. No finetuning is needed after AtomNAS finishes.}
    \item With the proposed search space and AtomNAS, we achieve state-of-the-art performance on ImageNet dataset under mobile setting.
\end{enumerate}

\section{Related Work}

\subsection{Neural Architecture Search}

Recently, there is a growing interest in automated neural architecture design. Reinforce learning based NAS methods \citep{zoph2017nasnet,tan2019mnasnet,tan2019mixnet,mingxing2019efficient} are usually computational intensive, thus hampering its usage with limited computational budget. To accelerate the search procedure, ENAS \citep{hieu2018enas} represents the search space using a directed acyclic graph and aims to search the optimal subgraph within the large supergraph. A training strategy of parameter sharing among subgraphs is proposed to significantly increase the searching efficiency. The similar idea of optimizing optimal subgraphs within a supergraph is also adopted by ~\cite{hanxiao2019darts,jin2019rc_darts,xu2020pc_darts,wu2019fbnet,zichao2019uniform_sampling,han2019proxyless}. \cite{stamoulis2019single_path,yu2020bignas} further share the parameters of different paths within a block using super-kernel representation.
% \prev{A prominent disadvantage of the above methods is their coarse search spaces only include limited categories of properties, e.g. kernel size, expansion ratio, the number of layer, etc. Because of the restriction of search space}
A prominent disadvantage of the above methods is that their coarse search spaces only support selecting one out of a set of choices (e.g., selecting one kernel size from \{3, 5, 7\}). MixNet tries to benefit from mixed operations by using a predefined set of mixed operations \{\{3\}, \{3, 5\}, \{3, 5, 7\}, \{3, 5, 7, 9\}\}, where the channels are equally distributed among different kernel sizes. Due to this limitation, it is difficult to learn optimal architectures under computational resource constraints. On the contrary, our method takes advantage of the fine-grained search space and is able to search for more flexible network architectures satisfying various resource constraints. The fine-grained search space proposed in this paper is exponentially larger than previous search space. For reference, the total number of possible structures within the experiment is around $10^{162}$, compared with $10^{21}$ for FBNet.
Recently, to improve the final performance of the searched architectures, \cite{yu2020bignas} utilizes knowledge distillation which is orthogonal to our method. It could be easily integrated into our method by Eq.~(\ref{eq:sparsity}) thanks to the end-to-end learning paradigm of our method.

\subsection{Network pruning}

%\jr{mention dynamic shrinkage could be used by network pruning method?}

Assuming that many parameters in the network are unnecessary, network pruning methods start from a computation-intensive model, identify the unimportant connections and remove them to get a compact and efficient network. Early method \citep{song2016deep_compression} simultaneously learns the important connections and weights. However, non-regularly removing connections in these works makes it hard to achieve theoretical speedup ratio on realistic hardwares due to extra overhead in caching and indexing. To tackle this problem, structured network pruning methods \citep{yihui2017channel,zhuang2017slimming,jianhao2017thinet,jianbo2018rethink_smallnorm,gordon2018morphnet} are proposed to prune structured components in networks, e.g. the entire channel and kernel. In this way, empirical acceleration can be achieved on modern computing devices.
\cite{zhuang2017slimming,jianbo2018rethink_smallnorm,gordon2018morphnet} encourage channel-level sparsity by imposing the L-1 regularizer on the channel dimension, which is also used by our method.
Recently, \cite{liu2019rethink_pruning} show that in structured network pruning, the learned weights are unimportant. This suggests structured network pruning is actually a neural architecture search focusing on channel numbers.
% This observation suggests that structured network pruning share common properties with neural architecture search as learning to prune channels can be seen as searching the optimal combination of channels from a search space consisting of independent channels.
Our method jointly searches the channel numbers and a mix of operations, which is a much larger search space.

% In our work, we challenge this point by our dynamic network shrinkage algorithm and proves the effectness of sparsity regularization.

\section{AtomNAS}

% TODO(meijieru): how to get here
We formulate our neural architecture search method in a fine-grained search space with the atomic block used as the basic search unit. An atomic block is comprised of two convolutions connected by a channel-wise operation. By stacking atomic blocks, we obtain larger building blocks (\textit{e.g.} residual block and MobileNetV2 block proposed in a variety of state-of-the-art models including ResNet, MobileNet V2/V3 \citep{he2016resnet,howard2019mobilenetv3,sandler2018mobilenetv2}. In Section~\ref{sec:fine_grained_search_space}, We first show larger network building blocks (\textit{e.g.} MobileNetV2 block) can be represented by an ensembles of atomic blocks. Based on this view, we propose a fine-grained search space using atomic blocks. In Section~\ref{sec:flops_targeted_slimming}, we propose a resource-aware atomic block selection method for end-to-end architecture search. Finally, we propose a dynamic network shrinkage technique in Section~\ref{sec:dynamic_network_shrinkage}, which greatly reduces the search cost.
% We start from reformulating the popular inverted residual block from the perspective of ensemble of multiple atom block in Sect.~\ref{sec:reformulate_bottleneck}. 

%Our method is based on the observation that many architectures contain a sub-structure: two convolutions joined by a channel-wise operation (\cite{he2016resnet,howard2019mobilenetv3,sandler2018mobilenetv2}); such sub-structure is also the search unit in some state-of-the-art NAS algorithms (\cite{tan2019mixnet,tan2019mnasnet}). In Section~\ref{sec:fine_grained_search_space}, We first provide an new perspective that regards such structure as an ensembles of multiple atom blocks; then we describe how this perspective leads us to extend the block-based search space into a more fine-grained one. Traditional NAS methods are not efficient when exploring this huge fine-grained search space. In Section~\ref{sec:flops_targeted_slimming}, we propose a \flops-targeted channel selection method as the solution. Finally, we propose a dynamic network shrinkage technique in Section~\ref{sec:dynamic_network_shrinkage}, which greatly reduces the searching cost.

\subsection{Fine-Grained Search Space}
\label{sec:fine_grained_search_space}

\begin{figure}[btp]
    \centering
    \includegraphics[width=1.0\textwidth]{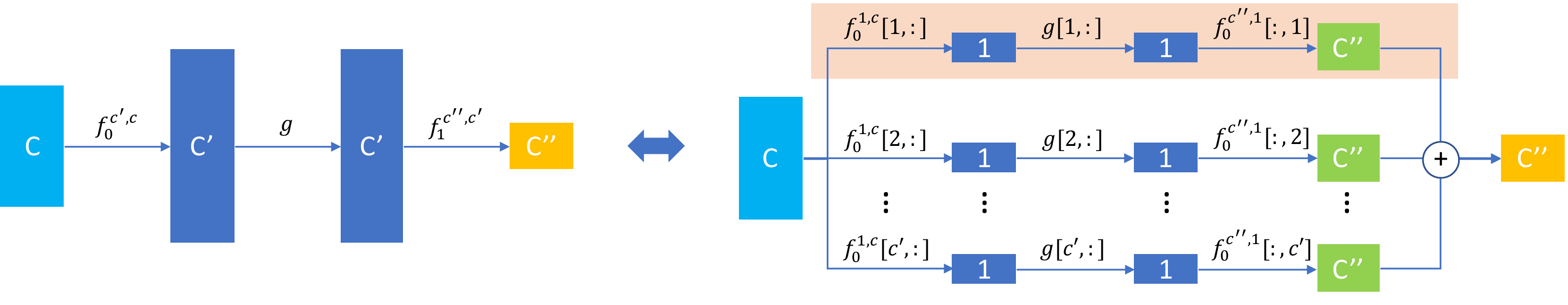}
    \caption{%\jr{change fig, illustrate vgg-like?}
    Illustration of the ensemble perspective. Arrow means operators. The structure of two convolutions joined by a channel-wise operation is mathematically equivalent to the ensemble of multiple atomic blocks, according to Eq.~(\ref{eq:atom_block}). Colored rectangles represent tensors, with numbers inside indicating their channel numbers; The shaded path on the right is one example of atomic block.} \label{fig:rethink_inverted_residual}
\end{figure}
% \begin{figure}[btp]
%     \begin{subfigure}[b]{\textwidth}
%         \centering
%         \includegraphics[width=0.47\textwidth]{figure/atomic_block2.pdf}
%         \caption{Ensemble of atomic blocks}
%         \label{fig:rethink_inverted_residual}
%     \end{subfigure}
%     \hfill
%     \begin{subfigure}[b]{0.47\textwidth}
%         \centering
%         \includegraphics[width=\textwidth]{figure/compound.pdf}
%         \caption{Fine-Grained Search Space}
%         \label{fig:compound_op}
%     \end{subfigure}
%     \caption{Illustration of the ensemble perspective and the proposed fine-grained search space. Arrow means operators. (a) The structure of two convolutions joined by a channel-wise operation is mathematically equivalent to the ensemble of multiple atomic blocks, according to Eq. (\ref{eq:atom_block}).
%     (b) The atomic block is the minimal unit. Here $g$ is a mix of three different operators, as indicated by the colors of the arrows; the third and the fifth atomic blocks are discarded. \jr{fix channel number or use another fig}
%     }
% \end{figure}
Under the typical block-wise NAS paradigm \citep{tan2019mnasnet,tan2019mixnet}, the search space of each block in a neural network is represented as the Cartesian product $\mathcal{C} = \prod_{i=1}\mathcal{P}_i$, where each $\mathcal{P}_i$ is the set of all choices of the $i$-th configuration such as kernel size, number of channels and type of operation. For example, $\mathcal{C}=\{\mbox{conv}, \mbox{depth-wise conv}, \mbox{dilated conv}\}\times \{3, 5\}\times\{24, 32, 64, 128\}$ represents a search space of three types of convolutions by two kernel sizes and four options of channel number. A block in the resulting model can only pick one convolution type from the three and one output channel number from the four values. This paradigm greatly limits the search space due to the few choices of each configuration. Here we present a more fine-grained search space by decomposing the network into smaller and more basic building blocks.
 %Under the typical block-wise paradigm (\cite{tan2019mnasnet,tan2019mixnet}), the search space of each block are represented as the Cartesian product $\mathcal{C} = \prod_{i=1}\mathcal{P}_i$, where each $\mathcal{P}_i$ is the set of all choices of the $i$-th property. Most existing NAS methods conduct search at this relatively coarse level. For example, $\mathcal{C}=\{\mbox{conv}, \mbox{depth-wise conv}, \mbox{dilated conv}\}\times \{3, 5\}\times\{24, 32, 64, 128\}$ represents a search space of three types of convolutions with two options of kernel size and four options of output channel number. A block can only picks one convolution type out of three and the output channel number can only be one of the four values. In this section, we first describe a new perspective of block search space and then derive a more fine-grained search space from it.

We denote $f^{c',c}(X)$ as a convolution operator, where $X$ is the input tensor and $c$, $c'$ are the input and output channel numbers respectively. A wide range of manually-designed and NAS architectures share a structure that joins two convolutions by a channel-wise operation:

%Denote by $f^{c',c}(X)$ a convolution operator, where $X$ is the input tensor and $c$, $c'$ are the input and output channel numbers respectively. The building blocks of many network architectures and the search units in many NAS algorithms have a structure that joins two convolutions by a channel-wise operator:
\begin{equation}
    Y = \left(f_1^{c'',c'} \circ g \circ f_0^{c',c}\right)(X) \label{eq:unit}
\end{equation}

%\jr{mention why could we deal with vgg like structure (use a subfig for illustration)}
where $g$ is a channel-wise operator. For example, in VGG \citep{simonyan2015vgg} and a Residual Block \citep{he2016resnet}, $f_0$ and $f_1$ are convolutions and $g$ is one of Maxpool, ReLU and BN-ReLU; in a MobileNetV2 block \citep{sandler2018mobilenetv2}, $f_0$ and $f_1$ are point-wise convolutions and $g$ is depth-wise convolution with BN-ReLU in the MobileNetV2 block.
%where $g$ is a channel-wise operator. For example, in VGG (\cite{simonyan2015vgg}) and ResNet (\cite{he2016resnet}), $f_0$ and $f_1$ are convolutions and $g$ is BN-ReLU; Inverted Residuals with Linear Bottleneck block (IRLB) in MobileNet V2 (\cite{sandler2018mobilenetv2}) has $f_0$ and $f_1$ as point-wise convolutions with $c''$ equal to $c$ and $g$ as BN-ReLU-depth-wise convolution-BN-ReLU; \cite{tan2019mixnet} use IRLB as their search unit.
Eq.~(\ref{eq:unit}) can be reformulated as follows:
\begin{equation}
    Y = \sum_{i=1}^{c'} \left(f_1^{c'',1}[i,:] \circ g[i,:] \circ f_0^{1,c}[:,i]\right)(X), \label{eq:atom_block}
\end{equation}
where $f_0^{1,c}[:,i]$ is the $i$-th convolution kernel of $f_0$, $g[i,:]$ is the operator of the $i$-th channel of $g$, and $\{f_1^{c'',1}[i,:]\}_{i=1}^{c'}$ are obtained by splitting the kernel tensor of $f_1$ along the the input channel dimension. Each term in the summation can be seen as a computationally independent block, which is called {\it atomic block}. Fig.~(\ref{fig:rethink_inverted_residual}) demonstrate this reformulation. By determining whether to keep each atomic block in the final model individually, the search of channel number $c'$ is enabled through channel selection, which greatly enlarges the search space. 

This formulation also naturally includes the selection of operators. To gain a better understanding, we first generalize Eq.~(\ref{eq:atom_block}) as:
\begin{equation}
    Y = \sum_{i=1}^{c'} \left(f_{1i}^{c'',1} \circ g_i \circ f_{0i}^{1,c}\right)(X). \label{eq:compound}
\end{equation}
Note the array indices $i$ are moved to subscripts. In this formulation, we can use different types of operators for $f_{0i}$, $f_{1i}$ and $g_i$; in other words, $f_0$, $f_1$ and $g$ can each be a combination of different operators and each atomic block can use different operators such as convolutions with different kernel sizes.

Formally, the search space is formulated as a supernet which is built based on the structure in Eq.~(\ref{eq:unit}); such structure satisfies Eq.~(\ref{eq:compound}) and thus can be represented by atomic blocks; each of $f_0$, $f_1$ and $g$ is a combination of operators. The new search space includes some state-of-the-art network architectures. For example, by allowing $g$ to be a combination of convolutions with different kernel sizes, the MixConv block in MixNet \citep{tan2019mixnet} becomes a special case in our search space. In addition, our search space facilitates discarding any number of channels in $g$, resulting in a more fine-grained channel configuration. In comparison, the channel numbers are determined heuristically in \cite{tan2019mixnet}.

\subsection{Resource-aware atomic block Search}
\label{sec:flops_targeted_slimming}

In this work, we adopt a differentiable neural architecture search paradigm where the model structure is discovered in a full pass of model training. With the supernet defined above, the final model can be produced by discarding part of the atomic blocks during training. Following DARTS (\cite{hanxiao2019darts}), we introduce a importance factor $\alpha$ to scale the output of each atomic block in the supernet. Eq.~(\ref{eq:compound}) then becomes
\begin{equation}
    Y = \sum_{i=1}^{c'} \alpha_i \left(f_{1i}^{c'',1} \circ g_i \circ f_{0i}^{1,c}\right)(X). \label{eq:weighted}
\end{equation}

Here, each $\alpha_i$ is tied with an atomic block comprised of three operators $f_{1i}^{c'',1}$,$g_i$ and $f_{0i}^{1,c}$. The importance factors are learned jointly with the network weights. Once the training finishes, the atomic blocks that have negligible effect (i.e., those with factors smaller than a threshold) on the network output are discarded.

We still need to address two issues related to the importance factors $\alpha_i$'s. The first issue is where in the supernet we should put the $\alpha$? Let's first consider the case when $g$ only contains linear operations, e.g., convolution, batch normalization and linear activation like ReLU. If $g$ contains at least one BN layer, The scaling parameters in the BN layers can be directly used as such importance factors (\cite{zhuang2017slimming}). If $g$ has no BN layers, which is rare, we can place $\alpha$ anywhere between $f_0$ and $f_1$; however, we need to apply regularization terms to the weights of $f_0$ and $f_1$ (e.g., weight decays) in order to prevent weights in $f_0$ and $f_1$ from getting too large and canceling the effect of $\alpha$. When $g$ contains non-linear operations, e.g., Swish activation and Sigmoid activation, we can only put $\alpha$ behind $f_1$.
%This approach is straightforward yet has two major issues. First, how to introduce these scaling factors into the supernet? In most cases, $g$ contains at least one BN layer; the only effective way then is to directly use scaling parameters of the last BN layer as $\alpha$, i.e., use the BN scale corresponding to the $i$-th channel as $\alpha_i$. Placing scaling factors before or after $g$ will make them being canceled by the BN layer. If $g$ has no BN layers, which is rare, theoretically we can place $\alpha$ anywhere between $f_0$ and $f_1$, as long as we put regularization on on the weights of $f_0$ and $f_1$ (e.g., weight decays); otherwise the effect of $\alpha$ will also be canceled.

The second issue is how to avoid performance deterioration after discarding some of the atomic blocks. For example, DARTS discards operations with small scale factors after iterative training of model parameters and scale factors. Since the scale factors of the discarded operations are not small enough, the performance of the network will be affected which needs re-training to adjust the weights again. In order to maintain the performance of the supernet after dropping some atomics blocks, the importance factors $\alpha$ of those atomic blocks should be sufficiently small. Inspired by the channel pruning work in \cite{zhuang2017slimming}, we add L1 norm penalty loss on $\alpha$, which effectively pushes many importance factors to near-zero values. At the end of learning, atomic blocks with $\alpha$ close to zero are removed from the supernet. Note that since the BN scales change more dramatically during training due to the regularization term, the running statistics of BNs might be inaccurate and needs to be calculated again using the training set.

With the added regularization term, the training loss is
\begin{equation}
    \mathcal{L} = \mathcal{E} + \lambda \sum_{i\in\mathcal{S}} c_i |\alpha_i|, \label{eq:sparsity}
\end{equation}
\begin{equation}
    c_i = \hat{c}_i / \sum_{k \in \mathcal{S}} \hat{c}_k
\end{equation}

where $\lambda$ is the coefficient of L1 penalty term, $\mathcal{S}$ is the index set of all atomic blocks, and $\mathcal{E}$ is the conventional training loss (e.g., cross-entropy loss combined with 
% \prev{the weight decay term}
the regularization term like weight decay and distillation loss.). $|\alpha_i|$ is weighted by coefficient $c_i$ which is proportional to the computation cost of $i$-th atomic block, i.e. $\hat{c}_i$. By using computation costs aware regularization, we encourage the model to learn network structures that strike a good balance between accuracy and efficiency.  In this paper, we use FLOPs as the criteria of computation cost. Other metrics such as latency and energy consumption can be used similarly. As a result, the whole loss function $\mathcal{L}$ trades off between accuracy and \flops.

\subsection{Dynamic Network Shrinkage}
\label{sec:dynamic_network_shrinkage}

\begin{figure}[t]
    \centering
    \includegraphics[width=0.5\linewidth]{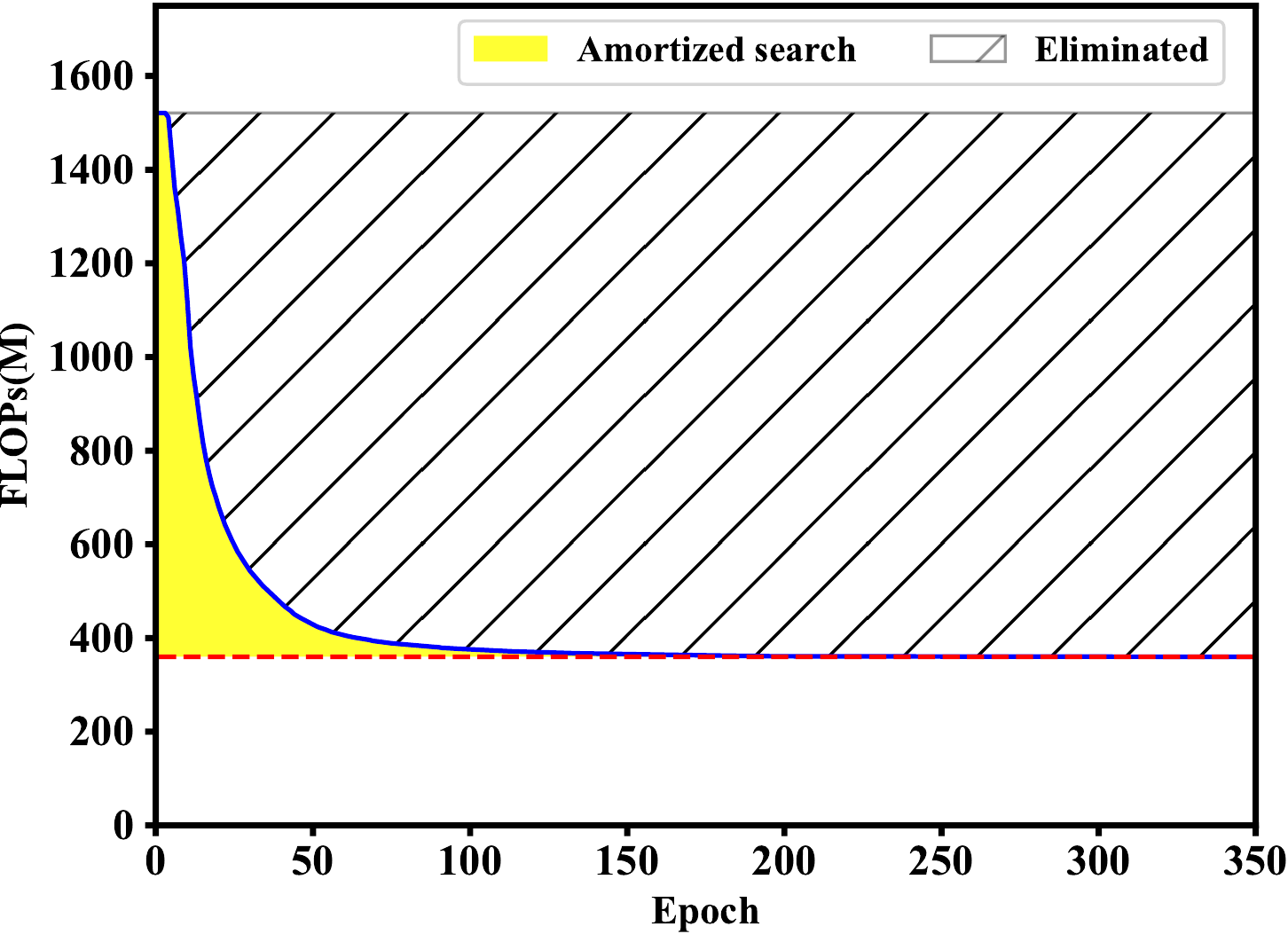}
    \caption{\flops change of the supernet during the searching and training for AtomNAS-C. The crossed-out region corresponds to the saved computation compared to training the supernet without the dynamic shrinkage. The region in yellow corresponds to the extra cost compared with training the final model from scratch, the cost of which is the region below the red dashed line.
    %It accelerates the search
    %We monitor the effective \flops by virtually mask out atom blocks with important factors less than $10^{-3}$, which found do not influence the prediction of the network. We highlight the \flops used for training the final model and amortized searching by red and yellow respectively. The ratio of these two areas are $17.2\%$.}
    }
    \label{fig:network_shrinkage}
\end{figure}

\begin{algorithm}[t]
    \SetAlgoLined
%    \KwResult{Searched architecture and its weight}
        Initialize the supernet\ and the exponential moving average\;
        \While{$epoch \le max\_epoch$}{
            Update network weights and importance factors $\alpha$ by minimizing the loss function $\mathcal{L}$ \;
            Update the $\hat{\alpha}$ by Eq.~(\ref{eq:alpha})\;
            \If{Total \flops of dead blocks $\ge \Delta$}
            {
                Remove dead blocks from the supernet\;
            }
            Recalculate BN's statistics by forwarding some training examples\;
            Validate the performance of the current supernet\;
        }
    \caption{Dynamic network shrinkage}
    \label{alg:dynamic_network_shrinkage}
\end{algorithm}

Usually, the supernet is much larger than the final search result. We observe that many atomic blocks become ``dead" starting from the early stage of the search, i.e., their importance factors $\alpha$ are close to zero till the end of the search. To utilize computational resources more efficiently and speed up the search process, we propose a dynamic network shrinkage algorithm which cuts down the network architecture by removing atomic blocks once they are deemed ``dead". 

We adopt a conservative strategy to decide whether an atomic block is ``dead": for importance factors $\alpha$, we maintain its momentum $\hat{\alpha}$ which is updated as
\begin{equation}
    \hat{\alpha} \leftarrow \beta \hat{\alpha} + (1-\beta)\alpha^t, \label{eq:alpha}
\end{equation}
where $\alpha^t$ is the importance factors at $t$-th iteration and $\beta$ is the decay term. An atomic block is considered ``dead" if both $\hat{\alpha}$ and $\alpha^t$ are smaller than a threshold, which is set to $0.001$ throughout experiments.

Once the total \flops of ``dead" blocks reach a predefined threshold, we remove those blocks from the supernet. As discussed above, we recalculate BN's running statistics before deploying the network. The whole training process is presented in Algorithm \ref{alg:dynamic_network_shrinkage}.

We show the \flops of a sample network during the search process in Fig.~\ref{fig:network_shrinkage}. We start from a supernet with 1521M \flops and dynamically discard ``dead" atomic blocks to reduce search cost. The overall search and train cost only increases by $17.2\%$ compared to that of training the searched model from scratch.
% \jr{mention cost for A/B?}

% TODO(meijieru): move to experiment
 %For AtomNAS-C model, with the help of the dynamic network shrinkage, the amortized search FLOPS is only $17.4\%$ compared with total FLOPS for training the searched architecture individually.

% $\hat{\gamma}_j$ is small means $\gamma_j$ is small in a sufficient long time.  Then we monitor the status of each atom block by comparing $\hat{\gamma}_j$ with a predefined small enough threshold $\epsilon$, which is $10^{-4}$ in our case.

\section{Experiment}

We first describe the implementation details in Section~\ref{sub:impl} and then compare AtomNAS with previous state-of-the-art methods under various \flops constraints in Section~\ref{sub:imagenet_expr}. 
% \prev{Finally}
In Section~\ref{sub:analysis}, we provide more detailed analysis about AtomNAS. Finally, in Section~\ref{sub:exp_coco}, we demonstrate the transferability of AtomNAS networks by evaluating them on detection and instance segmentation tasks.

% \jr{mention exp on coco, rename sections}

\subsection{Implementation Details}
\label{sub:impl}

\begin{figure}[t]
\centering
% \rule{5.4cm}{4.6cm}
\includegraphics[width=0.375\linewidth]{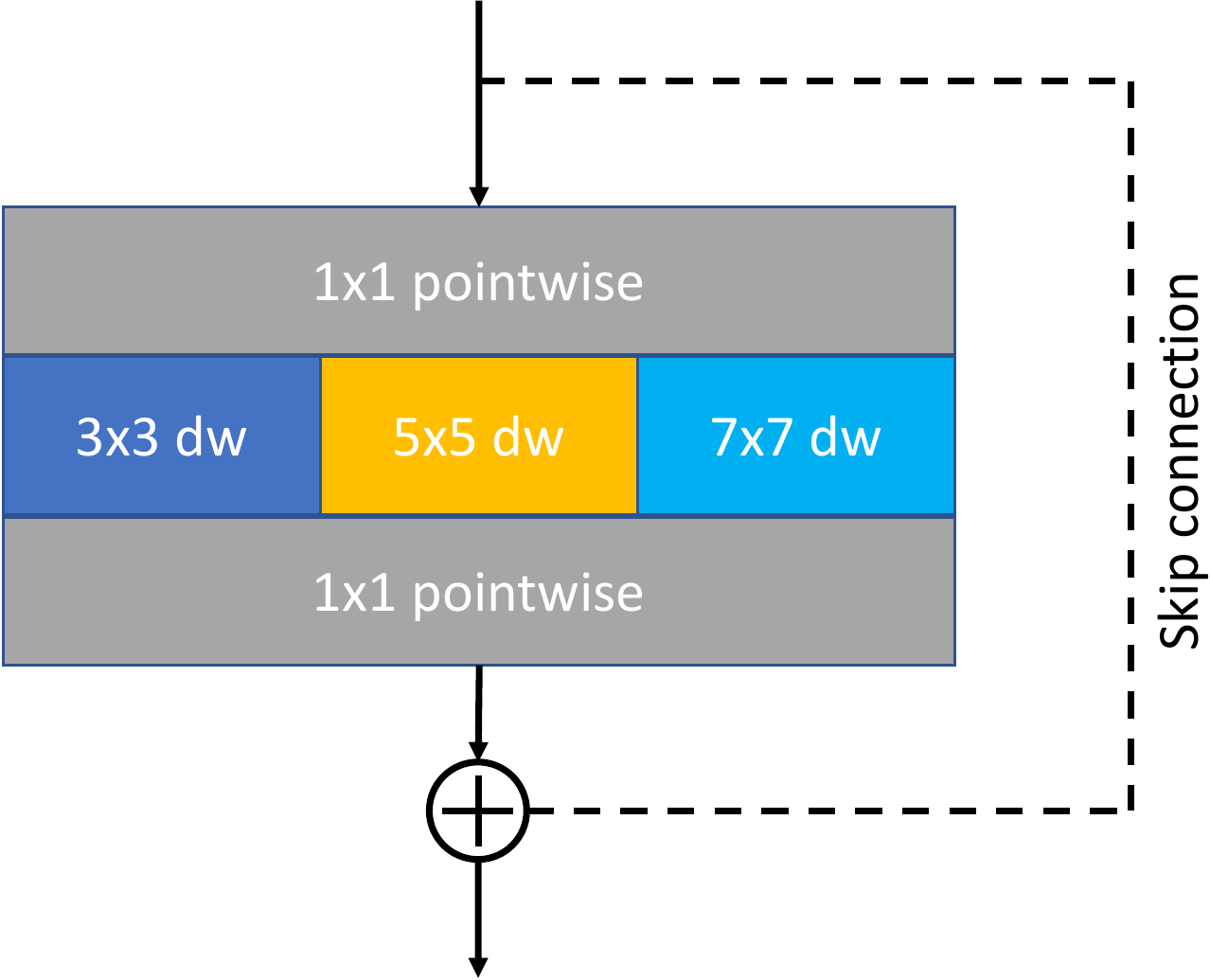}
\qquad
\begin{tabular}[b]{lllll}
\toprule
Input Shape & Block & f & n & stride \\
\midrule
$ 224^2 \times 3 $ & 3x3 conv & 32(16) & 1 & 2 \\
$ 112^2 \times 32(16) $ & 3x3 MB & 16 & 1 & 1 \\
$ 112^2 \times 16 $ & searchable & 24 & 4 & 2 \\
$ 56^2 \times 24 $ & searchable & 40 & 4 & 2 \\
$ 28^2 \times 40 $ & searchable & 80 & 4 & 2 \\
$ 14^2 \times 80 $ & searchable & 96 & 4 & 1 \\
$ 14^2 \times 96 $ & searchable & 192 & 4 & 2 \\
$ 7^2 \times 192 $ & searchable & 320 & 1 & 1 \\
$ 7^2 \times 320 $ & avgpool & - & 1 & 1 \\
$ 1280 $ & fc & 1000 & 1 & - \\
\bottomrule
\end{tabular}
% \captionlistentry[table]{}
% \label{tab:macro_arch}
% \captionsetup{labelformat=andtable}
\caption{(Left) The searchable block of the supernet. $f_0$ and $f_1$ are fixed to $1\times1$ pointwise convolutions; $g$ here is a mix of three convolutions with kernel sizes of $3\times3$, $5\times5$ and $7\times7$. $f_0$ expands the input channel number from $C$ to $18C$ and $f_1$ projects the channel number to the output channel number. If the output dimension
stays the same as the input dimension, we use a skip connection to add the input to the output. (Right) Architecture of the supernet. Column-Block denotes the block type; MB denotes MobileNetV2 block; "searchable" means a searchable block shown on the left.  Column-f denotes the output channel number of a block. Column-n denotes the number of blocks. Column-s denotes the stride of the first block in a stage. The output channel numbers of the first convolution are 16 for AtomNAS-A, 32 for AtomNAS-B and AtomNAS-C.}
\label{fig:tbs}
\end{figure}

The architecture of the supernet we use for the experiments is shown in table on the right of Fig. \ref{fig:tbs}. The supernet contains 21 AtomNAS blocks, the searchable block in our supernet; the picture on the right of Fig. \ref{fig:tbs} illustrates the structure of an AtomNAS block, where $f_0$ is a $1\times1$ pointwise convolutions that expands the input channel number from $C$ to $3\times6C$; $g$ is a mix of three depth-wise convolutions with kernel sizes of $3\times3$, $5\times5$ and $7\times7$, and $f_1$ is another $1\times1$ pointwise convolutions that projects the channel number to the output channel number. Similar to MobileNetV2~\citep{sandler2018mobilenetv2}, if the output dimension stays the same as the input dimension, we use a skip connection to add the input to the output. AtomNAS block is effectively an ensemble of $3\times6C$ atomic blocks, whose underlying search space covers the MobileNetV2 block~\citep{sandler2018mobilenetv2} and its multi-kernel variant, MixConv~\citep{tan2019mixnet}. Within AtomNAS block, we are able to optimize the distribution of computation resources (i.e., channel numbers) among the three types of depth-wise convolution.

% In our experiments, we use MobileNetV2 block as the building block, as it is widely adopted in many existing NAS algorithms (!!!cite!!!). (!!! use figure and table!!!) The supernet uses the backbone structure of \cite{stamoulis2019single_path}. For each layer with input channel $C$, we have $6 \cdot C$ candidates atomic blocks for each kernel size from $\{3, 5, 7\}$. Totally we have 21 searchable layers, and in each layer, there are $(6C)^3$ choices. Thus the total number of possible structures within the fine-grained search space is $2.8 \times 10^{162}$, which is orderly larger than previous search space. For reference, the number of possible architectures in FBNet \citep{wu2019fbnet} is around $10^{21}$. Powered by our network shrinkage algorithm, we could directly search and train the architecture simultaneously in such huge search space on the complete ImageNet training set without any proxy.

We use the same training configuration (e.g., RMSProp optimizer, EMA on weights and exponential learning rate decay) as \cite{tan2019mnasnet,stamoulis2019single_path} and do not use extra data augmentation such as MixUp \citep{zhang2018mixup} and AutoAugment \citep{cubuk2018autoaugment}. We find that using this configuration is sufficient for our method to achieve good performance. Our results are shown in Table~\ref{tab:overall_compare} and Table~\ref{tab:bn_calib}. When training the supernet, we use a total batch size of 2048 on 32 Tesla V100 GPUs and train for 350 epochs. For our dynamic network shrinkage algorithm, we set the momentum factor $\beta$ in Eq.~(\ref{eq:alpha}) to $0.9999$. At the beginning of the training, all of the weights are randomly initialized. To avoid removing atomic blocks with high penalties (i.e., \flops) prematurely, the weight of the penalty term in Eq.~(\ref{eq:sparsity}) is increased from 0 to the target $\lambda$ by a linear scheduler during the first 25 epochs. By setting the weight of the L1 penalty term $\lambda$ to be $1.8 \times 10^{-4}$, $1.2\times 10^{-4}$ and $1.0 \times 10^{-4}$ respectively, we obtain networks with three different sizes: AtomNAS-A, AtomNAS-B, and AtomNAS-C. They have the similar \flops as previous state-of-the-art networks under $400$M: MixNet-S \citep{tan2019mixnet}, MixNet-M \citep{tan2019mixnet} and SinglePath \citep{stamoulis2019single_path}. In Appendix \ref{sec:vis}, we visualize the architecture of AtomNAS-C.

% \yingwei{Table~\ref{tab:macro_arch} and Fig.~\ref{fig:tbs} can be correctly refed.}

% In general, . When search-and-train, we use a batch size of 2048 as the supernet requiring more GPU memory. We will give an solution for this issue in Section~\ref{ssub:search_cost} powered by our dynamic network shrinkage algorithm.

\subsection{Experiments on ImageNet}
\label{sub:imagenet_expr}

We apply AtomNAS to search high performance light-weight model on ImageNet 2012 classification task \citep{deng2009imagenet}. Table~\ref{tab:overall_compare} compares our methods with previous state-of-the-art models, either manually designed or searched. 

With models directly produced by AtomNAS, our method achieves the new state-of-the-art under all \flops constraints. Especially, AtomNAS-C achieves $75.9\%$ top-1 accuracy with only 360M \flops, and surpasses all other models, including models like PDARTS and DenseNAS which have much higher \flops.

% Especially, with the same but fine-grained search space, our AtomNAS outperforms previous state-of-the-art SinglePathNAS by $0.5\%$ with less \flops, which reveals the importance of fine-grained search space and end-to-end searching. 

Techniques like Swish activation function \citep{ramachandran2018swish} and Squeeze-and-Excitation (SE) module \citep{hu2018se} consistently improve the accuracy with marginal \flops cost. For a fair comparison with methods that use these techniques, we directly modify the searched network by replacing all ReLU activation with Swish and add SE module with ratio 0.5 to every block and then retrain the network from scratch. Note that unlike other methods, we do not search the configuration of Swish and SE, and therefore the performance might not be optimal. Extra data augmentations such as MixUp and AutoAugment are still not used. We train the models from scratch with a total batch size of 4096 on 32 Tesla V100 GPUs for 250 epochs. 
% It's worth to note that due to the regularization by sparsity, simply retrain always result in inferior accuracy, which is explained later. We increase the dropout rate into $0.28$ for compensation a bit.

Simply adding these techniques improves the results further. AtomNAS-A+ achieves $76.3\%$ top-1 accuracy with 260M \flops, which outperforms many heavier models including MnasNet-A2. Without extra data augmentations, it performs as well as Efficient-B0 \citep{mingxing2019efficient} by using 130M less \flops. It also outperforms the previous state-of-the-art MixNet-S by $0.5\%$. In addition, AtomNAS-C+ improves the top-1 accuracy on ImageNet to $77.6\%$, surpassing previous state-of-the-art MixNet-M by $0.6\%$ and becomes the overall best performing model under 400M \flops.
% AtomNAS-L+ achieves $79.2\%$ top-1 accuracy, which is the overall state-of-the-art for mobile setting under 600M.
% We plot the result in Fig.~\ref{fig:overall_imagenet}.

\begin{figure}[t]
    \centering
    \includegraphics[width=0.7\linewidth]{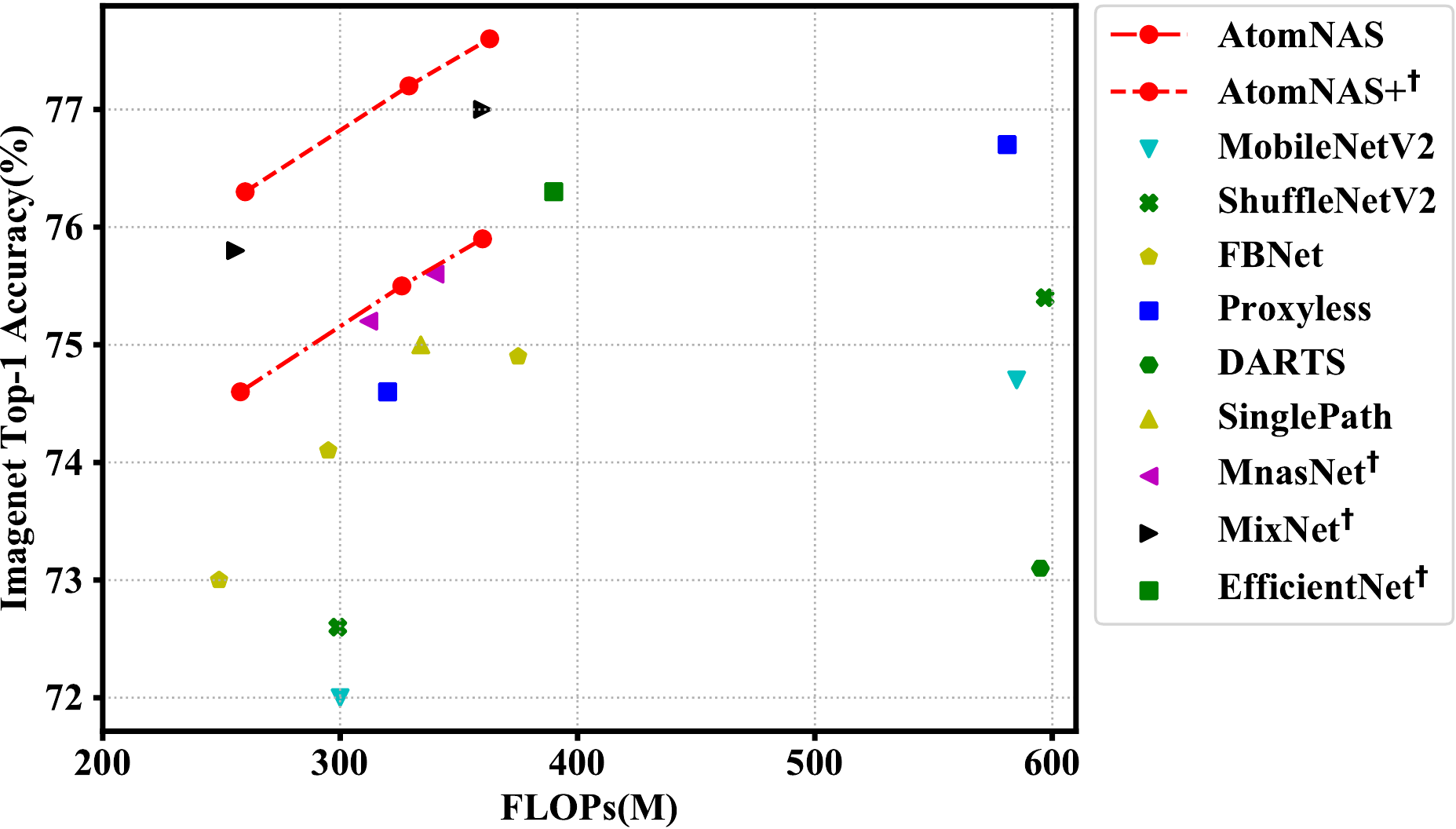}
    \caption{\flops versus accuracy on ImageNet. $^\dag$ means methods use extra techniques like Swish activation and Squeeze-and-Excitation module.}
    \label{fig:overall_imagenet}
\end{figure}

Fig.~\ref{fig:overall_imagenet} visualizes the top-1 accuracy on ImageNet for different models. It's clear that our fine-grained search space and the end-to-end resource-aware search method boost the performance significantly.

\begin{table}[t]
\caption{Comparision with state-of-the-arts on ImageNet under the mobile setting. $^\dag$ denotes methods using extra network modules such as Swish activation and Squeeze-and-Excitation module. $^\ddag$ denotes using extra data augmentation such as MixUp and AutoAugment. $^*$ denotes models searched and trained simultaneously.}
\label{sample-table}
\begin{center}
\begin{tabular}{lllll}
\toprule
Model & Parameters & \flops & Top-1(\%) & Top-5(\%) \\
\midrule
MobileNetV1 \citep{howard2017mobilenet}           & 4.2M & 575M & 70.6 & 89.5 \\
MobileNetV2 \citep{sandler2018mobilenetv2}        & 3.4M & 300M & 72.0 & 91.0 \\
MobileNetV2 (our impl.)                           & 3.4M & 301M & 73.6 & 91.5 \\
MobileNetV2 (1.4) & 6.9M & 585M & 74.7 & 92.5 \\
ShuffleNetV2 \citep{ma2018shufflenet_v2}          & 3.5M & 299M & 72.6 & - \\
ShuffleNetV2 2$\times$      & 7.4M & 591M & 74.9 & - \\
% CondenseNet (G=C=8) \citep{huang2018condensenet}  & 2.9M & 274M & 71.0 & 90.0 \\
% CondenseNet (G=C=4)  & 4.8M & 529M & 73.8 & 91.7 \\
\midrule
% MnasNet \citep{tan2019mnasnet}  & 4.2M & 317M & 74.0 & 91.8 \\
FBNet-A \citep{wu2019fbnet}         & 4.3M & 249M & 73.0 & - \\
% FBNet-B \citep{wu2019fbnet}         & 4.5M & 295M & 74.1 & - \\
FBNet-C         & 5.5M & 375M & 74.9 & - \\
Proxyless (mobile) \citep{han2019proxyless}                  & 4.1M & 320M & 74.6 & 92.2 \\
% Proxyless (GPU)                   & 7.1M & 465M & 75.1 & 92.5 \\
SinglePath \citep{stamoulis2019single_path} & 4.4M & 334M & 75.0 & 92.2 \\
NASNet-A \citep{zoph2017nasnet}     & 5.3M & 564M & 74.0 & 91.6 \\
DARTS (second order) \citep{hanxiao2019darts}      & 4.9M & 595M & 73.1 & - \\
PDARTS (cifar 10) \citep{chen2019pdarts}       & 4.9M & 557M & 75.6 & 92.6 \\
DenseNAS-A \citep{fang2019densenas} & 7.9M & 501M & 75.9 & 92.6 \\
% DARTS+ (Cifar 100) \citep{liang2019dartsplus} & 5.1M & 591M & 76.3 & 92.8 \\
% OneShot \footnote{Latest result from official implementation: \url{https://github.com/megvii-model/ShuffleNet-Series/tree/master/OneShot}} (block + channel) \citep{zichao2019uniform_sampling} & 3.4M & 319M & 74.9 & 92.0 \\
FairNAS-A \citep{chu2019fairnas} & 4.6M & 388M & 75.3 & 92.4 \\

\textbf{AtomNAS-A}$^*$  & 3.9M & 258M & 74.6 & 92.1 \\
\textbf{AtomNAS-B}$^*$  & 4.4M & 326M & 75.5 & 92.6 \\
\textbf{AtomNAS-C}$^*$  & 4.7M & 360M & 75.9 & 92.7 \\

\midrule

% SinglePath-SE$^\dag$ \citep{stamoulis2019single_path_automl} &
SCARLET-A$^\dag$ \citep{chu2019scarletnas} & 6.7M & 365M & 76.9 & 93.4 \\
MnasNet-A1$^\dag$ \citep{tan2019mnasnet}      & 3.9M & 312M & 75.2 & 92.5 \\
MnasNet-A2$^\dag$      & 4.8M & 340M & 75.6 & 92.7 \\
MixNet-S$^\dag$ \citep{tan2019mixnet}      & 4.1M & 256M & 75.8 & 92.8 \\
MixNet-M$^\dag$      & 5.0M & 360M & 77.0 & 93.3 \\
EfficientNet-B0$^{\dag \ddag}$ \citep{mingxing2019efficient} & 5.3M & 390M & 76.3 & 93.2 \\
SE-DARTS+$^{\dag \ddag}$ \citep{liang2019dartsplus} & 6.1M & 594M & 77.5 & 93.6 \\
% Mixnet-L$^\dag$      & 7.3M & 565M & 78.9 & 94.2 \\

% TODO(meijieru): uniform sampling and other
% \midrule
% AtomNAS-L  & 7.2M & 564M & - & - \\
\textbf{AtomNAS-A+}$^\dag$ & 4.7M & 260M & 76.3 & 93.0 \\
\textbf{AtomNAS-B+}$^\dag$ & 5.5M & 329M & 77.2 & 93.5 \\
\textbf{AtomNAS-C+}$^\dag$ & 5.9M & 363M & 77.6 & 93.6 \\
% AtomNAS-L$^\dag$ & 9.1M & 568M & 79.0 & 94.4 \\
\bottomrule
\end{tabular}
\end{center}
\label{tab:overall_compare}
\end{table}

\subsection{Analysis}    
\label{sub:analysis}

\subsubsection{Resource-Aware Regularization}

To demonstrate the effectiveness of the resource-aware regularization in Section ~\ref{sec:flops_targeted_slimming}, we compare it with a baseline without \flops-related coefficients $c_i$, which is widely used in network pruning \citep{zhuang2017slimming,yihui2017channel}.
Table~\ref{tab:equal_baseline} shows the results. First, by using the same L1 penalty coefficient $\lambda=1.0 \times 10^{-4}$, the baseline achieves a network with similar performance but using much more \flops; then by increasing $\lambda$ to $1.5 \times 10^{-4}$, the baseline obtain a network which has similar \flops but inferior performance (i.e., about $1.0\%$ lower). In Fig.~\ref{fig:channel_pie_equal} we visualized the ratio of different types of atomic blocks of the baseline network obtained by $\lambda=1.5 \times 10^{-4}$. The baseline network keeps more atomic blocks in the earlier blocks, which have higher computation cost due to higher input resolution. On the contrary, AtomNAS is aware of the resource constraint, thus keeping more atomic blocks in the later blocks and achieving much better performance.

\begin{table}[htb]
\caption{Influence of awareness of resource metric. The upper block uses equal penalties for all atomic blocks. The lower part uses our resource-aware atomic block selection.}
\begin{center}
\begin{tabular}{lll}
\toprule
$\lambda$ & \flops & Top-1(\%) \\
\midrule
$1.0 \times 10^{-4}$ & 445M & 76.1 \\
$1.5 \times 10^{-4}$ & 370M & 74.9 \\
\midrule
$1.0 \times 10^{-4}$ & 360M & 75.9 \\
\bottomrule
\end{tabular}
\end{center}
\label{tab:equal_baseline}
\end{table}

\subsubsection{BN Recalibration}

% As for our dynamic network shrinkage algorithm, naively discarding the dead atom blocks may destroy the network, as the influence of the bias terms of them is non-trivial for the running mean of the following BNs. On the other hand, due to the regularization on the BN's scales, they changes relatively rapidly than without regularization. It results in instability for the running variances approximation within the following BNs, which deteriorates the result significantly.
As the BN's running statistics might be inaccurate as explained in Section~\ref{sec:flops_targeted_slimming} and Section~\ref{sec:dynamic_network_shrinkage}, we re-calculate the running statistics of BN before inference, by forwarding 131k randomly sampled training images through the network. Table~\ref{tab:bn_calib} shows the impact of the BN recalibration. The top-1 accuracies of AtomNAS-A, AtomNAS-B, and AtomNAS-C on ImageNet improve by $1.4\%$, $1.7\%$, and $1.2\%$ respectively, which clearly shows the benefit of BN recalibration. 
% We also reimplement several models and do the calibration. First, the MnasNet training scheduler consistently boost the performance for all model. Second, the improvement from calibration are much smaller than ours.

% For exploring the influence of BN calibration, we rerun the experiment without dynamic network shrinkage to get model AtomNAS-C1, AtomNAS-M2 and AtomNAS-A1 by setting $\lambda$ to $1.0^{-4}, 1.2^{-4}$ and $1.5^{-4}$ respectively, and compare the results in Table \ref{tab:bn_calib}. For calibration, we use 81920 training images with batch size 256 per GPU, which could be done in just few seconds. After calibration, the accuracy boosts $1.4\%$ from $74.8\%$ to $76.2\%$. Also, for fair comparison, we re-implement some state-of-the-arts and calibrate the running statistics. Calibration consistently improves the performances though the improvements are much smaller than ours.

\begin{table}[htb]
\caption{Influence of BN recalibration.}
\begin{center}
\begin{tabular}{lccc}
\toprule
Model   & w/o Recalibration & w/ Recalibration\\
\midrule
% M1$\dag$ & 358M & - & 74.8 & 76.2 (+1.4) \\
% M2$\dag$ & 323M & - & 74.0 & 75.7 (+1.7) \\
% S1$\dag$ & 293M & - & 73.5 & 75.2 (+1.7) \\
% \midrule
% M1$\ddag$ & 358M & - & 74.8 & 76.2 (+1.4) \\
AtomNAS-A  & 73.2 & 74.6 (+1.4) \\
AtomNAS-B  & 73.8 & 75.5 (+1.7) \\
AtomNAS-C  & 74.7 & 75.9 (+1.2) \\
% \midrule
% SinglePath & 334M & 75.0$^*$ & 75.0 & 75.1 (+0.1) \\
% Proxyless (mobile) & 320M & 74.6 & 74.9 & 75.1 (+0.2) \\
% MobileNetV2 & 300M & 72.0 & 73.5 & 73.7 (+0.2) \\
\bottomrule
\end{tabular}
\end{center}
\label{tab:bn_calib}
\end{table}

\subsubsection{Cost of Dynamic Network Shrinkage}
\label{ssub:search_cost}

Our dynamic network shrinkage algorithm speedups the search and train process significantly. For AtomNAS-C, the total time for search-and-training is 25.5 hours. For reference, training the final architecture from scratch takes 22 hours.
% Thus the amortized search cost is just 4 hours when .
Note that as the supernet shrinks, both the GPU memory consumption and forward-backward time are significantly reduced. Thus it's possible to dynamically change the batch size once having sufficient GPU memory, which would further speed up the whole procedure.

\subsection{Experiments on COCO Detection and Instance Segmentation}
\label{sub:exp_coco}

In this section, we assess the performance of AtomNAS models as feature extractors for object detection and instance segmentation on COCO dataset~\citep{lin2014mscoco}. We first pretrain AtomNAS models (without Swish activation function \citep{ramachandran2018swish} and Squeeze-and-Excitation (SE) module \citep{hu2018se}) on ImageNet, use them as drop-in replacements for the backbone in the Mask-RCNN model~\citep{he2017maskrcnn} by building the detection head on top of the last feature map, and finetune the model on COCO dataset.

We use the open-source code MMDetection~\citep{kai2019mmdetection}. All the models are trained on COCO train2017 with batch size 16 and evaluated on COCO val2017. Following the schedule used in the open-source implementation of TPU-trained
Mask-RCNN \footnote{\url{https://github.com/tensorflow/tpu/tree/master/models/official/mask_rcnn}}, the learning rate starts at $0.02$ and decreases by a scale of 10 at 15-th and 20th epoch respectively. The models are trained for 23 epochs in total.

Table~\ref{tab:detection} compares the results with other baseline backbone models. The detection results of baseline models are from \cite{stamoulis2019single_path_automl}. We can see that all three AtomNAS models outperform the baselines on object detection task. The results demonstrate that our models have better transferability than the baselines, which may due to mixed operations, a.k.a multi-scale here, are more important to object detection and instance segmentation.

\begin{table}[htbp]
\caption{Comparision with baseline backbones on COCO object detection and instance segmentation. Cls denotes the ImageNet top-1 accuracy; detect-mAP and seg-mAP denotes mean average precision for detection and instance segmentation on COCO dataset. The results of baseline models are from \cite{stamoulis2019single_path_automl}. SinglePath+~\citep{stamoulis2019single_path_automl} contains SE module.}
\begin{center}
\begin{tabular}{lllll}
% \begin{tabular}{llllllll}
\toprule
% Model & \flops & ImageNet Top-1\% & bbox-mAP & bbox-mAP$_S$ & bbox-mAP$_M$ & bbox-mAP$_L$ & seg-mAP & seg-mAP$_S$ & seg-mAP$_M$ & seg-mAP$_L$ &  \\
Model & \flops & Cls (\%) & detect-mAP (\%) & seg-mAP (\%)  \\
\midrule
%MobileNetV2 (our impl.)                           & 301M & 73.6 & 29.7 & 27.8 \\
MobileNetV2 \citep{sandler2018mobilenetv2}                           & 301M & 73.6 & 30.5 & - \\
% MobileNetV2 \jr{(our impl.)}                           & 301M & 73.6 & 30.5 & - \\
\midrule
Proxyless (mobile) \citep{han2019proxyless}                  & 320M & 74.6 & 32.9 & - \\
Proxyless (mobile) (our impl.)                  & 320M & 74.9 & 32.7 & 30.0 \\
SinglePath+ \citep{stamoulis2019single_path_automl} & 353M & 75.6 & 33.0 & - \\
SinglePath (our impl.) & 334M & 75.0 & 32.0 & 29.7 \\
\midrule
\textbf{AtomNAS-A}  & 258M & 74.6 & 32.7 & 30.1 \\
\textbf{AtomNAS-B}  & 326M & 75.5 & 33.6 & 30.8 \\
\textbf{AtomNAS-C}  & 360M & 75.9 & 34.1 & 31.4 \\
% \midrule
% EfficientNet-B0 (our impl.)  & 390M & 77.0 & 32.8 & 30.4 \\
\bottomrule

\end{tabular}
\end{center}
\label{tab:detection}
\end{table}

% \subsubsection{Relationship with Network Pruning}
% 
% Pruning usually start from a network designed by experts which is relatively compact compared with the supernet in our method. \citep{luo2018autopruner} prunes during finetune.
% % Also, traditional network pruning method
% 
% TODO(meijieru): discuss with pruning
% - why converge faster
%     - start from a redundant supernet
%     - not pretrained
%     - remove **dead** blocks only

\section{Conclusion}
In this paper, we revisit the common structure, i.e., two convolutions joined by a channel-wise operation, and reformulate it as an ensemble of atomic blocks. This perspective enables a much larger and more fine-grained search space. For efficiently exploring the huge fine-grained search space, we propose an end-to-end framework named AtomNAS, which conducts architecture search and network training jointly. The searched networks achieve significantly better accuracy than previous state-of-the-art methods while using small extra cost.

%\new{Some future work include exploring the influence of different pruning methods within the proposed framework. Another interesting topic is to search the topology of network instead of predefining it.}
%Experiments 

% Currently, our methods reuses BN's scale factor for channel selections, thus only apply to layers with BN. How to extend the fine-grained search space to layers without BN, like pooling, skip connection, is an important direction worth to explore.

\bibliography{iclr2020_conference}
\bibliographystyle{iclr2020_conference}

\appendix
\section{Visualization}\label{sec:vis}

\begin{figure}[htbp]
    \centering
    \includegraphics[width=\linewidth]{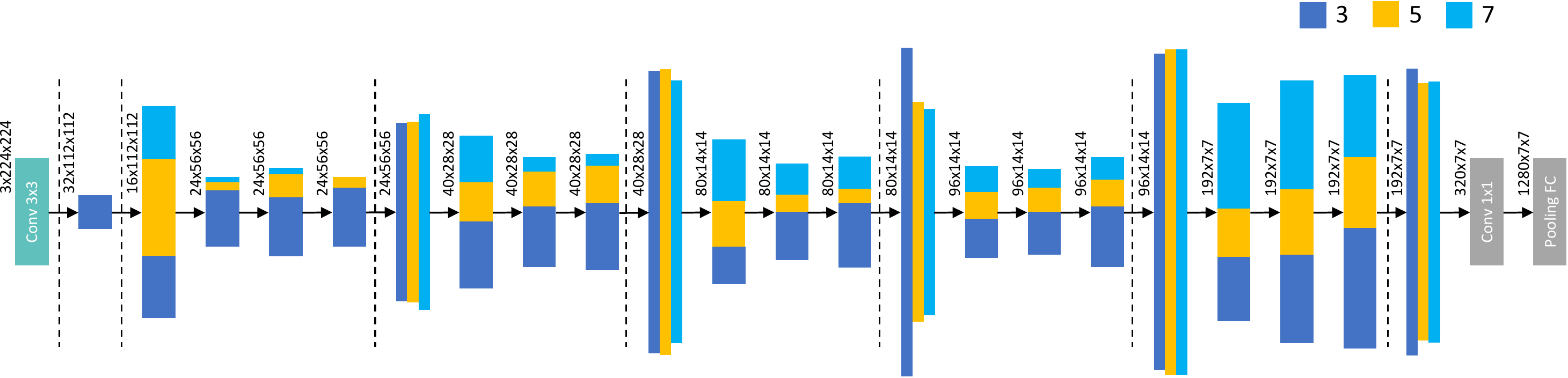}
    \caption{The architecture of AtomNAS-C. Blue, orange, cyan blocks denote atomic blocks with kernel size $3$, $5$ and $7$ respectively; the heights of these blocks are proportional to their expand ratios.}
    \label{fig:atomnas_m_arch}
\end{figure}

\begin{figure}[btp]
    \begin{subfigure}[b]{\textwidth}
        \centering
        \includegraphics[width=\textwidth]{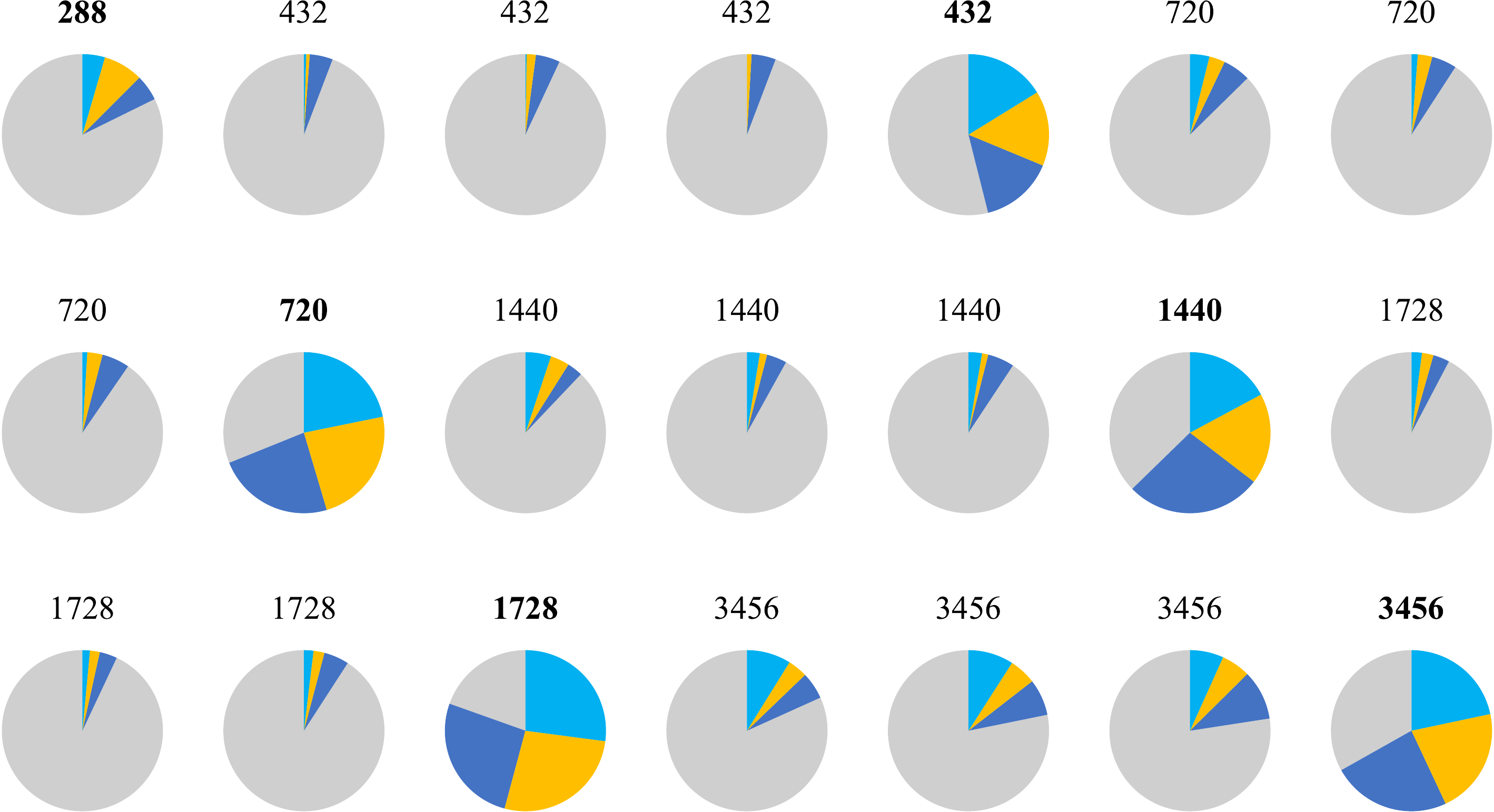}
        \caption{AtomNAS-C}
        \label{fig:mmodel_pie}
    \end{subfigure}
    % \vfill
    \\[3ex]
    \begin{subfigure}[b]{\textwidth}
        \centering
        \includegraphics[width=\textwidth]{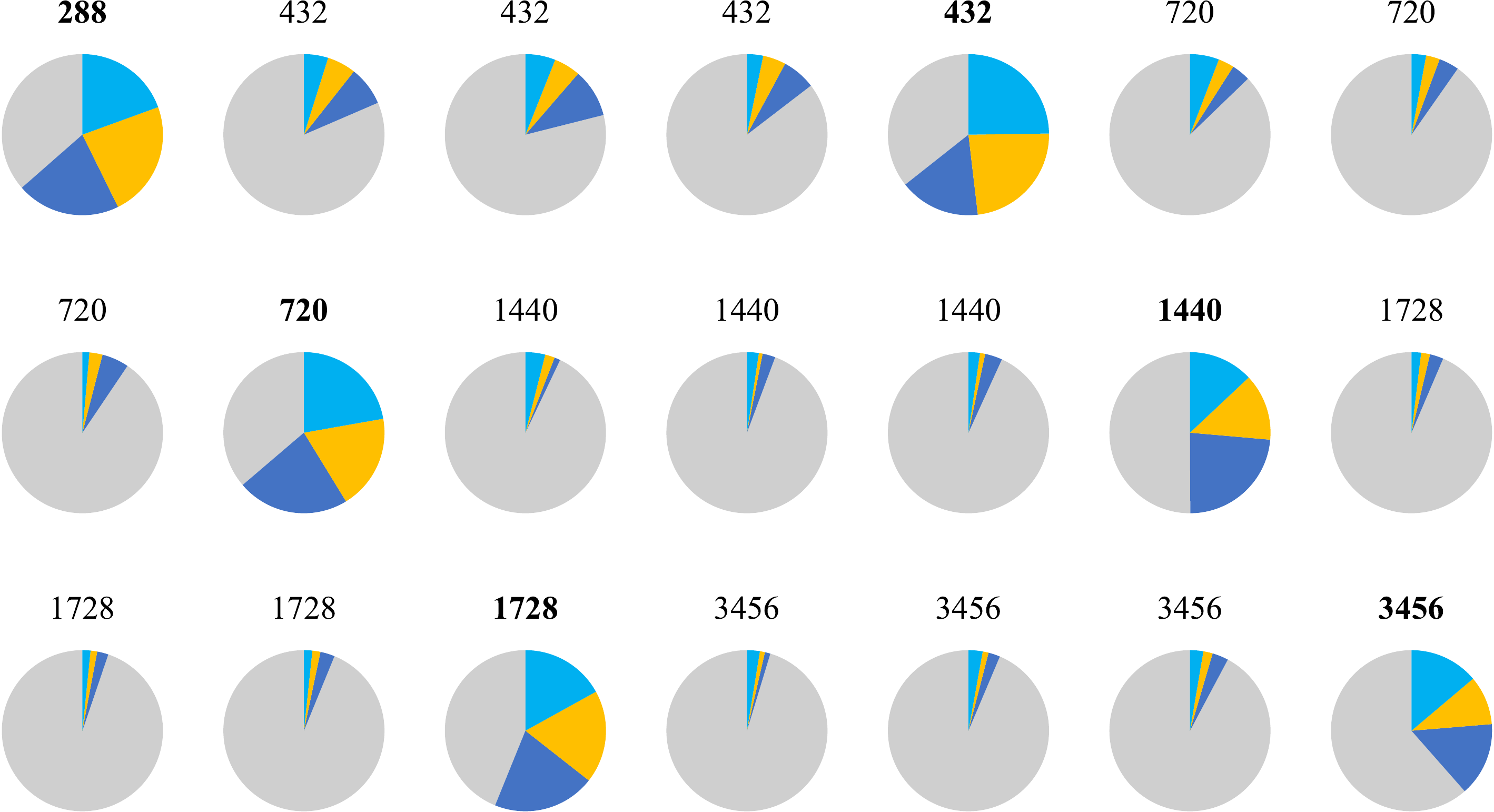}
        \caption{Baseline}
        \label{fig:channel_pie_equal}
    \end{subfigure}
    \caption{Ratio of different types of atomic blocks in all 21 searchable blocks.
    The text above each pie tells the total number of atomic blocks of the corresponding block in the original supernet. Grey denotes dead atomic blocks; blue, orange, and cyan represent atomic blocks using depth-wise convolutions with kernel size $3,5,7$ respectively. Blocks without skip connection are highlighted by bold text.
    (a) Visualization for AtomNAS-C. (b) Visualization for baseline (i.e., without \flops related coefficients $c_i$).
    }
\end{figure}

% \begin{figure}[htbp]
%     \centering
%     \includegraphics[width=\linewidth]{figure/architecture-crop.pdf}
%     \caption{The architecture of AtomNAS-C. Blue, orange, cyan blocks denote atomic blocks with kernel size $3$, $5$ and $7$ respectively; the heights of these blocks are proportional to their expand ratios.}
%     \label{fig:atomnas_m_arch}
% \end{figure}

We plot the structure of the searched architecture AtomNAS-C in Fig.~\ref{fig:atomnas_m_arch}, from which we see more flexibility of channel number selection, not only among different operators within each block, but also across the network. In Fig.~\ref{fig:mmodel_pie}, we visualize the ratio between atomic blocks with different kernel sizes in all 21 search blocks. First, we notice that all search blocks have convolutions of all three kernel sizes, showing that AtomNAS learns the importance of using multiple kernel sizes in network architecture. Another observation is that AtomNAS tends to keep more atomic blocks at the later stage of the network. This is because in earlier stage, convolutions of the same kernel size costs more \flops; AtomNAS is aware of this (thanks to its resource-aware regularization) and try to keep as less as possible computationally costly atomic blocks.

\end{document}

%% file: math_commands.tex
\usepackage{amsmath,amsfonts,bm}

\def\eqref#1{equation~\ref{#1}}
\def\1{\bm{1}}

\DeclareMathAlphabet{\mathsfit}{\encodingdefault}{\sfdefault}{m}{sl}
\SetMathAlphabet{\mathsfit}{bold}{\encodingdefault}{\sfdefault}{bx}{n}